\documentclass[11pt,a4paper]{article}
\pdfoutput=1
\usepackage[hyperref]{acl2021}
\usepackage{times}
\usepackage{latexsym}

\usepackage{hyperref}
\usepackage{url}

\usepackage{graphicx} 
\usepackage{subfig}
\graphicspath{{figs/}}

\usepackage{array}
\usepackage{tabularx}
\usepackage{makecell}
\usepackage{multirow}
\usepackage{booktabs}
\usepackage[para]{threeparttable}

\usepackage{comment}
\usepackage{xspace}

\usepackage{amsfonts}
\usepackage{mathtools}
\DeclarePairedDelimiter{\ceil}{\lceil}{\rceil}

\usepackage{microtype}

\aclfinalcopy 
\def\aclpaperid{4210} 

\newcommand{\model}{Length-Adaptive Transformer\xspace}
\newcommand{\dr}{Drop-and-Restore\xspace}
\newcommand{\ld}{LengthDrop\xspace}

\newcommand{\lab}{Length-Adaptive$^\star$}
\newcommand{\las}{Length-Adaptive$^\dagger$}

\newcommand{\pb}{PoWER-BERT\xspace}
\newcommand{\db}{DistilBERT\xspace}
\newcommand{\bb}{BERT\textsubscript{Base}\xspace}

\title{\model: \\Train Once with Length Drop, Use Anytime with Search}

\author{
Gyuwan Kim \\ Clova AI, NAVER Corp. \\ \texttt{gyuwan.kim@navercorp.com} \\
\And
Kyunghyun Cho \\ New York University \\ \texttt{kyunghyun.cho@nyu.edu} \\
}

\begin{document}

\maketitle

\begin{abstract}
Despite transformers' impressive accuracy, their computational cost is often prohibitive to use with limited computational resources. Most previous approaches to improve inference efficiency require a separate model for each possible computational budget. In this paper, we extend PoWER-BERT \citep{goyal2020power} and propose \textit{\model} that can be used for various inference scenarios after one-shot training. We train a transformer with \textit{\ld}, a structural variant of dropout, which stochastically determines a sequence length at each layer. We then conduct a multi-objective evolutionary search to find a length configuration that maximizes the accuracy and minimizes the efficiency metric under any given computational budget. Additionally, we significantly extend the applicability of PoWER-BERT beyond sequence-level classification into token-level classification with \textit{\dr} process that drops word-vectors temporarily in intermediate layers and restores at the last layer if necessary. We empirically verify the utility of the proposed approach by demonstrating the superior accuracy-efficiency trade-off under various setups, including span-based question answering and text classification.
\ifaclfinal
Code is available at \href{https://github.com/clovaai/length-adaptive-transformer}{https://github.com/clovaai/length-adaptive-transformer}.
\else
Source code will be available upon publication.
\fi

\end{abstract}

\section{Introduction}
\label{sec:introduction}

Pre-trained language models~\citep{peters2018deep, devlin2018bert, radford2019language, yang2019xlnet, he2020deberta} have achieved notable improvements in various natural language processing (NLP) tasks.
Most of them rely on transformers~\citep{vaswani2017attention}, and the number of model parameters ranges from hundreds of millions to billions~\citep{shoeybi2019megatron, raffel2019exploring, kaplan2020scaling, brown2020language}.
Despite this high accuracy, excessive computational overhead during inference, both in terms of time and memory, has hindered its use in real applications.
This level of excessive computation has further raised the concern over energy consumption as well~\citep{schwartz2019green, strubell2019energy, cao2020towards}.

Recent studies have attempted to address these concerns regarding large-scale transformers' computational and energy efficiency (see \S\ref{sec:related_work} for a more extensive discussion.) Among these, we focus on PoWER-BERT~\citep{goyal2020power} which progressively reduces sequence length by eliminating word-vectors based on the attention values as passing layers.
PoWER-BERT establishes the superiority of accuracy-time trade-off over earlier approaches~\citep{sanh2019distilbert,sun2019patient,michel2019sixteen}.
However, it requires us to train a separate model for each efficiency constraint. 
In this paper, we thus develop a framework based on PoWER-BERT such that we can train a single model that can be adapted in the inference time to meet any given efficiency target.

In order to train a transformer to cope with a diverse set of computational budgets in the inference time, we propose to train once while reducing the sequence length with a random proportion at each layer. We refer to this procedure as \ld, which was motivated by the nested dropout~\citep{rippel2014learning}.
We can extract sub-models of shared weights with any length configuration without requiring extra post-processing nor additional fine-tuning.

It is not trivial to find an optimal length configuration given the inference-time computational budget, although it is extremely important in order to deploy these large-scale transformers in practice. 
Once a transformer is trained with the proposed \ld, we search for the length configuration that maximizes the accuracy given a computational budget. Because this search is combinatorial and has multiple objectives (accuracy and efficiency), in this work, we propose to use an evolutionary search algorithm, which further allows us to obtain a full Pareto frontier of accuracy-efficiency trade-off of each model. 

PoWER-BERT, which forms the foundation of the proposed two-stage procedure, is only applicable to sequence-level classification, because it eliminates some of the word vectors at each layer by design.
In other words, it cannot be used for token-level tasks such as span-based question answering \citep{rajpurkar2016squad} because these tasks require hidden representations of the entire input sequence at the final layer.
We thus propose to extend PoWER-BERT with a novel \dr{} process (\textsection \ref{sec:drop-and-restore-proceess}), which eliminates this inherent limitation.  
Word vectors are dropped and set aside, rather than eliminated, in intermediate layers to maintain the saving of computational cost, as was with the original PoWER-BERT. 
These set-aside vectors are then restored at the final hidden layer and provided as an input to a subsequent task-specific layer, unlike the original PoWER-BERT.

The main contributions of this work are two-fold. First, we introduce \ld, a structured variant of dropout for training a single \model{} model that allows us to automatically derive multiple sub-models with different length configurations in the inference time using evolutionary search, without requiring any re-training. Second, we design \dr{} process that makes PoWER-BERT applicable beyond classification, which enables PoWER-BERT to be applicable to a wider range of NLP tasks such as span-based question answering.  We empirically verify \model{} works quite well using the variants of BERT on a diverse set of NLP tasks, including SQuAD 1.1~\citep{rajpurkar2016squad} and two sequence-level classification tasks in GLUE benchmark~\citep{wang2018glue}. 
Our experiments reveal that the proposed approach grants us fine-grained control of computational efficiency and a superior accuracy-efficiency trade-off in the inference time compared to existing approaches.

\section{Background}  
\label{sec:background}

In this section, we review some of the building blocks of our main approach. In particular, we review transformers, which are a standard backbone used in natural language processing these days, and PoWER-BERT, which was recently proposed as an effective way to train a large-scale, but highly efficient transformer for sequence-level classification.

\subsection{Transformers and BERT}
\label{sec:transformers_and_bert}

A transformer is a particular neural network that has been designed to work with a variable-length sequence input and is implemented as a stack of self-attention and fully connected layers~\citep{vaswani2017attention}. 
It has recently become one of the most widely used models for natural language processing. 
Here, we give a brief overview of the transformer which is the basic building block of the proposed approach.

Each token $x_t$ in a sequence of tokens $x = (x_1, \dots, x_N)$, representing input text, is first turned into a continuous vector $h^0_t \in \mathbb{R}^{H}$ which is the sum of the token and position embedding vectors. 
This sequence is fed into the first transformer layer which returns another sequence of the same length $h^1 \in \mathbb{R}^{N \times H}$. 
We repeat this procedure $L$ times, for a transformer with $L$ layers, to obtain $h^L = (h_1^L, \ldots, h_N^L)$. 
We refer to each vector in the hidden sequence at each layer as a {\it word vector} to emphasize that there exists a correspondence between each such vector and one of the input words.

Although the transformer was first introduced for the problem of machine translation, \citet{devlin2018bert} demonstrated that the transformer can be trained and used as a sentence encoder. More specifically, \citet{devlin2018bert} showed that the transformer-based masked language model, called BERT, learns a universally useful parameter set that can be fine-tuned for any downstream task, including sequence-level and token-level classification.

In the case of sequence-level classification, a softmax classifier is attached to the word vector $h^L_1$ associated with the special token \texttt{[CLS]}, and the entire network, including the softmax classifier and BERT, is fine-tuned. For token-level classification, we use each $h^L_t$ as the final hidden representation of the associated $t$-th word in the input sequence.
This strategy of pre-training followed by fine-tuning, often referred to as transfer learning, is a dominant approach to classification in natural language processing.

\subsection{PoWER-BERT}
\label{sec:power-bert}

PoWER-BERT keeps only the topmost $l_j$ word vectors at each layer $j$ by eliminating redundant ones based on the significance score which is the total amount of attention imposed by a word on the other words \citep{goyal2020power}.
$l_j$ is the hyper-parameter that determines how many vectors to keep at layer $j$.
PoWER-BERT has the same model parameters as BERT, but the extraction layers are interspersed after the self-attention layer in every transformer block \citep{vaswani2017attention}.

PoWER-BERT reduces inference time successfully, achieving better accuracy-time trade-off than DistilBERT \citep{sanh2019distilbert}, BERT-PKD \citep{sun2019patient}, and Head-Prune \citep{michel2019sixteen}.
Despite the original intention of maximizing the inference efficiency with the minimal loss in accuracy, it is possible to set up PoWER-BERT to be both more efficient and more accurate compared to the original BERT,
which was observed but largely overlooked by \citet{goyal2020power}.

Training a PoWER-BERT model consists of three steps: (1) fine-tuning, (2) length configuration search, and (3) re-training.
The fine-tuning step is just like the standard fine-tuning step of BERT given a target task.
A length configuration is a sequence of retention parameters $(l_1, \cdots l_{L})$, each of which corresponds to the number of word vectors that are kept at each layer. These retention parameters are learned along with all the other parameters to minimize the original task loss together with an extra term that approximately measures the number of retained word vectors across layers.
In the re-training step, PoWER-BERT is fine-tuned with the length configuration fixed to its learned one.

For each computational budget, we must train a separate model going through all three steps described above. 
Moreover, the length configuration search step above is only approximate, as it relies on the relaxation of retention parameters which are inherently discrete. This leads to the lack of guaranteed correlation between the success of this stage and true run-time.
Even worse, it is a delicate act to tune the length configuration given a target computational budget because the trade-off is {\it implicitly} made via a regularization coefficient. 
Furthermore, PoWER-BERT has an inherent limitation in that it only applies to sequence-level classification because it eliminates word vectors in intermediate layers.

\section{\model}

In this section, we explain our proposed framework which results in a transformer that reduces the length of a sequence at each layer with an arbitrary rate. We call such a resulting transformer a \model{}. 
We train \model{} with \ld{} which randomly samples the number of hidden vectors to be dropped at each layer with the goal of making the final model robust to such drop in the inference time.
Once the model is trained, we search for the optimal trade-off between accuracy and efficiency using multi-objective evolutionary search, which allows us to use the model for any given computational budget without fine-tuning nor re-training. 
At the end of this section, we describe \dr{} process as a way to greatly increase the applicability of PoWER-BERT which forms a building block of the proposed framework. 

In short, we train a \model{} once with \ld{} and \dr{}, and use it with an automatically determined length configuration for inference with any target computational budget, on both sequence-level and token-level tasks.

\subsection{\ld}  
\label{sec:lengthdrop}

Earlier approaches to efficient inference with transformers have focused on a scenario where the target computational budget for inference is known in advance \cite{sanh2019distilbert, goyal2020power}. This greatly increases the cost of deploying transformers, as it requires us to train a separate transformer for each scenario. Instead, we propose to train one model that could be used for a diverse set of target computational budgets without re-training. 

Before each SGD update, \ld{} randomly generates a length configuration by sequentially sampling a sequence length $l_{i + 1}$ at the $(i+1)$-th layer based on the previous layer's sequence length $l_i$, following the uniform distribution  $\mathcal{U}(\ceil{(1 - p)l_i}, l_i)$, where $l_0$ is set to the length of the input sequence, and $p$ is the \ld{} probability.
This sequential sampling results in a length configuration $(l_1, \cdots, l_L)$.
\model{} can be thought of as consisting of a full model and many sub-models corresponding to different length configurations, similarly to a neural network trained with different dropout masks \citep{srivastava2014dropout}.

\paragraph{LayerDrop}

From the perspective of each word vector, the proposed \ld{} could be thought of as skipping the layers between when it was set aside and the final layer where it was restored. The word vector however does not have any information based on which it can determine whether it would be dropped at any particular layer. In our preliminary experiments, we found that this greatly hinders optimization. We address this issue by using LayerDrop~\citep{fan2019reducing} which skips each layer of a transformer uniformly at random. The LayerDrop encourages each word vector to be agnostic to skipping any number of layers between when it is dropped and when it is restored, just like dropout~\citep{srivastava2014dropout} prevents hidden neurons from co-adapting with each other by randomly dropping them.

\paragraph{Sandwich Rule and Inplace Distillation} 

We observed that standard supervised training with \ld does not work well in the preliminary experiments.
We instead borrow a pair of training techniques developed by \citet{yu2019universally} which are sandwich rule and inplace distillation, for better optimization as well as final generalization.
At each update, we update the full model without \ld{} as usual to minimize the supervised loss function. We simultaneously update $n_s$ randomly-sampled sub-models (which are called sandwiches) and the smallest-possible sub-model, which corresponds to 
keeping only $\ceil{(1-p)l_i}$ word vectors at each layer $i$, using knowledge distillation~\citep{hinton2015distilling} from the full model.
Here, sub-models mean models with length reduction. They are trained to their prediction close to the full model's prediction (inplace distillation).

\subsection{Evolutionary Search of Length Configurations}
\label{sec:evolutionary-search}

After training a \model{} with \ld{}, we search for appropriate length configurations for possible target computational budgets that will be given at inference time. The length configuration determines the model performance in terms of both accuracy and efficiency.
In order to search for the optimal length configuration, we propose to use evolutionary search, similarly to \citet{cai2019once} and \citet{wang2020hat}. This procedure is efficient, as it only requires a single pass through the relatively small validation set for each length configuration, unlike re-training for a new computational budget which requires multiple passes through a significantly larger training set for each budget. 

We initialize the population with constant-ratio configurations. Each configuration is created by $l_{i + 1} = \ceil{(1 - r)l_i}$ for each layer $i$ with $r$ so that the amount of computation within the initial population is uniformly distributed between those of the smallest and full models.
At each iteration, we evolve the population to consist only of configurations lie on a newly updated efficiency-accuracy Pareto frontier by mutation and crossover. 
Mutation alters an original length configuration $(l_1, \cdots, l_L)$ to $(l'_1, \cdots, l'_L)$ by sampling $l'_i$ from the uniform distribution $\mathcal{U}(l'_{i-1}, l_{i+1})$ with the probability $p_m$ or keeping the original length $l'_i = l_i$, sweeping the layers from $i = 1$ to $i = L$.
A crossover takes two length configurations and averages the lengths at each layer.
Both of these operations are performed while ensuring the monotonicity of the lengths over the layers. 
We repeat this iteration $G$ times while maintaining $n_m$ mutated configurations and $n_c$ crossover'd configurations.
Repeating this procedure pushes the Pareto frontier further to identify the best trade-off between two objectives, efficiency and accuracy, without requiring any continuous relaxation of length configurations nor using a proxy objective function.

\subsection{\dr~Process}
\label{sec:drop-and-restore-proceess}

\begin{figure*}[t!] 
    \centering

    \subfloat[\textit{PoWER}]{\includegraphics[width=0.44\textwidth]{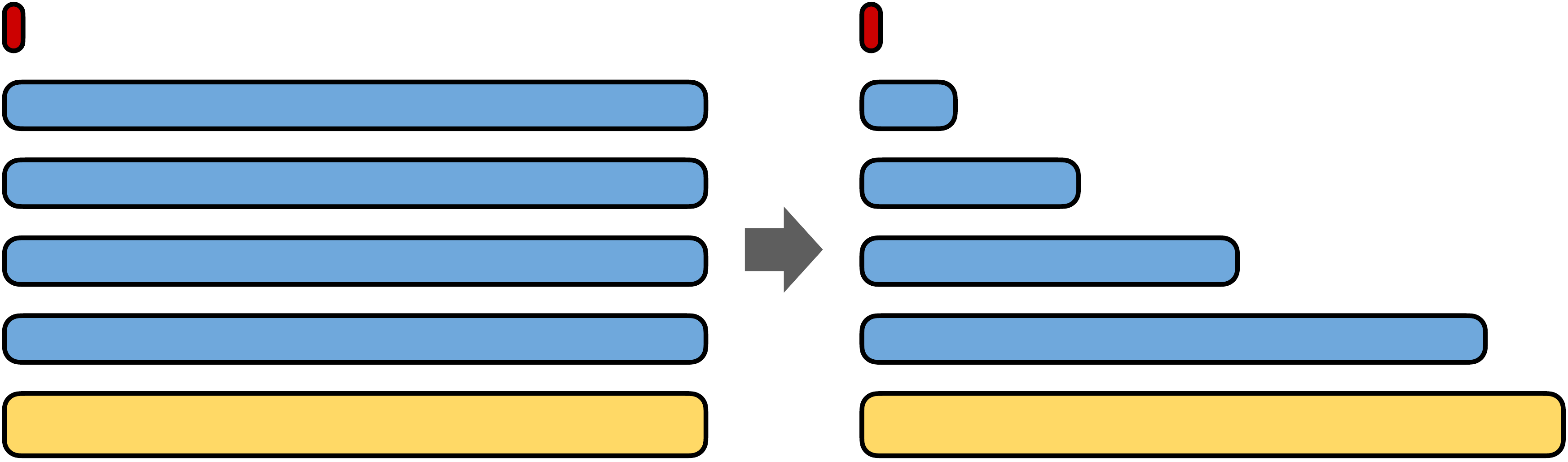}}
    \qquad
    \subfloat[\textit{\dr}]{\includegraphics[width=0.44\textwidth]{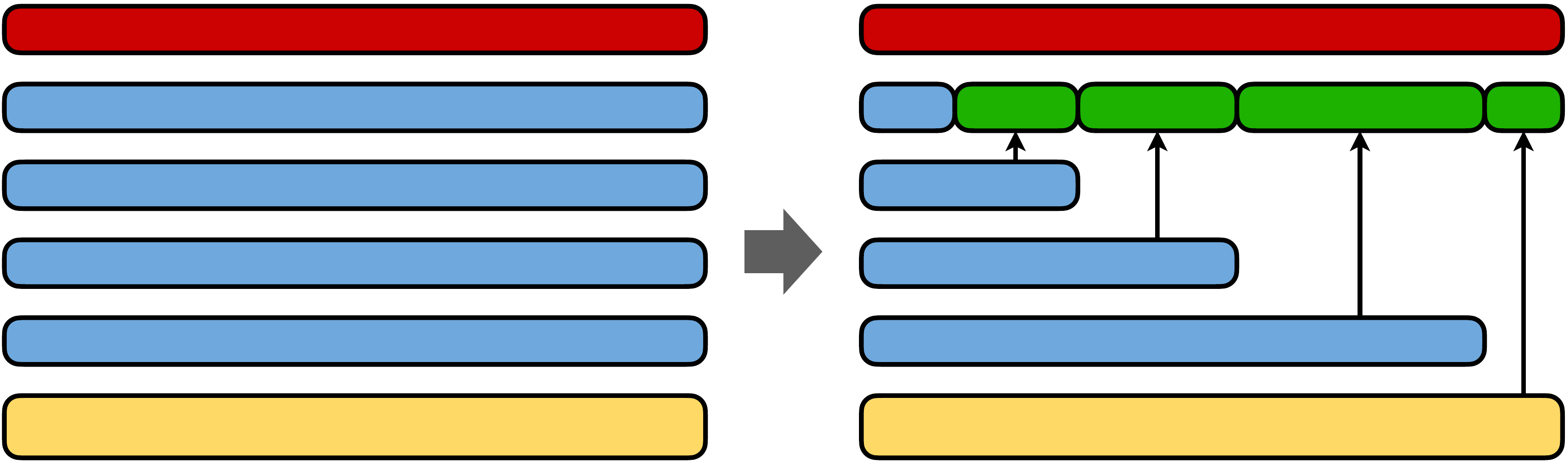}}

    \caption{
    Illustration of (a) word-vector elimination process in PoWER-BERT \citep{goyal2020power} and (b) \dr~process in \model. 
    Yellow box and blue boxes imply the output of embedding layer and transformer layers, respectively.
    Green boxes mean vectors dropped in lower layers and restored at the last layer. 
    Red box is the task-specific layer.
    Though word-vectors in the middle could be eliminated (or dropped), remaining vectors are left-aligned for the better illustration.
    In this case, the number of transformer layers is four.
    }
    \label{fig:drop-and-restore}
\end{figure*}

The applicability of the PoWER-BERT, based on which our main contribution above was made, is limited to sequence-level classification because it eliminates word vectors at each layer. In addition to our main contribution above, we thus propose to extend the PoWER-BERT so that it is applicable to token-level classification, such as span-based question-answering. Our proposal, to which we refer as \dr{}, does not eliminate word vectors at each layer according to the length configuration but instead sets them aside until the final hidden layer. At the final hidden layer, these word vectors are brought back to form the full hidden sequence, as illustrated graphically in Figure~\ref{fig:drop-and-restore}.

\section{Experiment Setup}

\paragraph{Datasets}

We test the proposed approach on both sequence-level and token-level tasks, the latter of which could not have been done with the original PoWER-BERT unless for the proposed \dr{}. We use MNLI-m and SST-2 from GLUE benchmark~\citep{wang2018glue}, as was done to test \pb{} earlier, for sequence-level classification. We choose them because consistent accuracy scores from standard training on them due to their sufficiently large training set imply that they are reliable to verify our approach.
We use SQuAD~1.1~\citep{rajpurkar2016squad} for token-level classification.

\paragraph{Evaluation metrics}

We use the number of floating operations (FLOPs) as a main metric to measure the inference efficiency given any length configuration, as it is agnostic to the choice of the underlying hardware, unlike other alternatives such as hardware-aware latency \citep{wang2020hat} or energy consumption \citep{henderson2020towards}. We later demonstrate that FLOPs and wall-clock time on GPU and CPU correlate well with the proposed approach, which is not necessarily the case for other approaches, such as unstructured weight pruning~\citep{han2015deep,see2016compression}.

\paragraph{Pre-trained transformers}

Since BERT was introduced by \citet{devlin2018bert}, it has become a standard practice to start from a pre-trained (masked) language model and fine-tune it for each downstream task. We follow the same strategy in this paper and test two pre-trained transformer-based language models; 
\bb \citep{devlin2018bert}
and DistilBERT \citep{sanh2019distilbert}, 
which allows us to demonstrate that the usefulness and applicability of our approach are not tied to any specific architectural choice, such as the number of layers and the maximum input sequence length.
Although we focus on BERT-based masked language models here, the proposed approach is readily applicable to any transformer-based models.

\paragraph{Learning}

We train a \model{} with \ld{} probability and LayerDrop probability both set to $0.2$. We use $n_s=2$ randomly sampled intermediate sub-models in addition to the full model and smallest model for applying the sandwich learning rule. 

We start fine-tuning the pre-trained transformer without \dr{} first, just as \citet{goyal2020power} did with PoWER-BERT. We then continue fine-tuning it for another five epochs {\it with} \dr{}. This is unlike the recommended three epochs by \citet{devlin2018bert}, as learning progresses slower due to a higher level of stochasticity introduced by \ld{} and LayerDrop. 
We use the batch size of $32$, the learning rate of $5e-5$ for SQuAD 1.1 and $2e-5$ for MNLI-m and SST, and the maximum sequence length of 384 for SQuAD 1.1 and 128 for MNLI-m and SST.

\paragraph{Search}

We run up to $G = 30$ iterations of evolutionary search, using $n_m = 30$ mutated configurations with mutation probability $p_m = 0.5$ and $n_c = 30$ crossover'd configurations, to find the Pareto frontier of accuracy and efficiency.

\begin{figure*}[!t]
\centering
\input{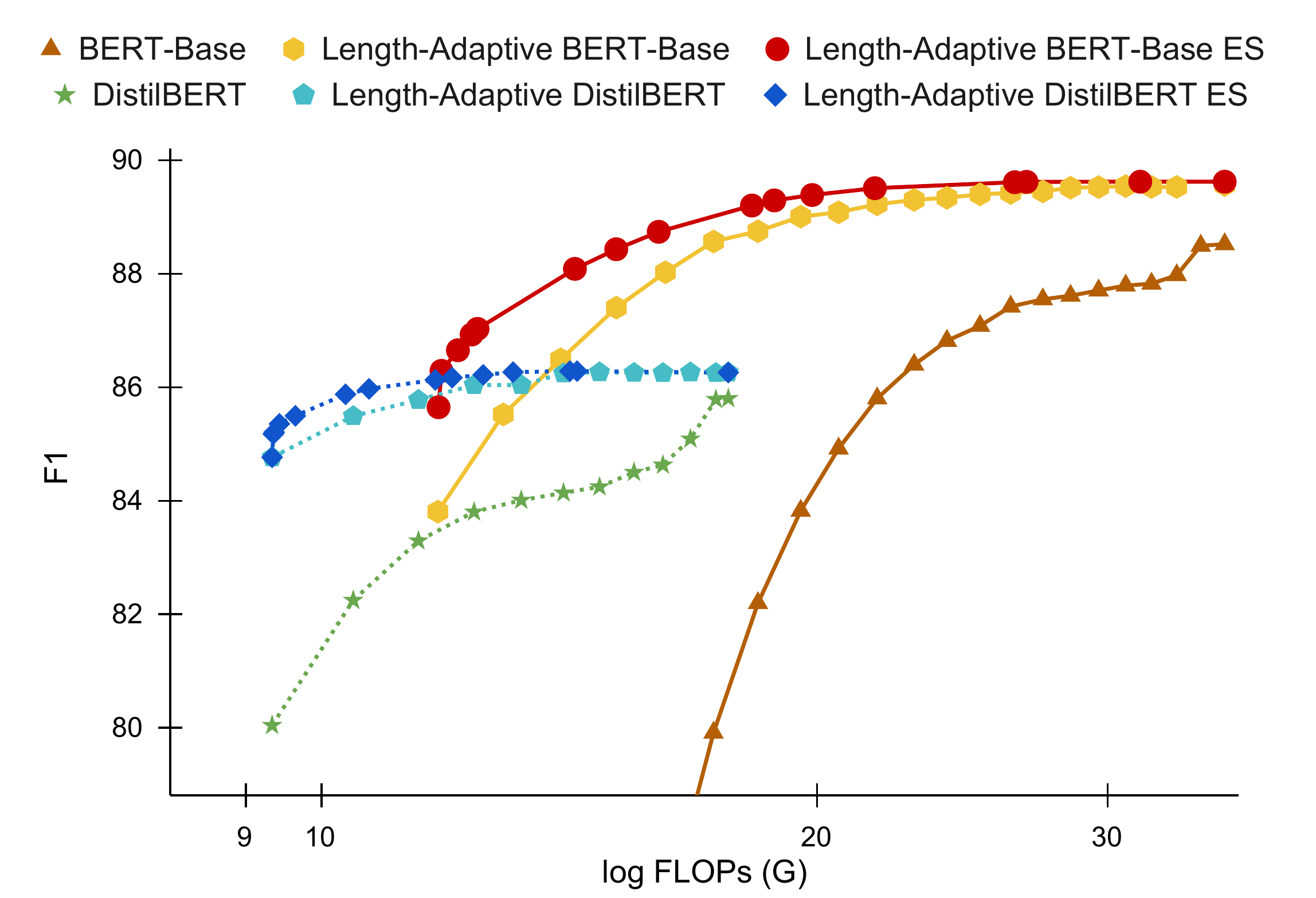}
\quad
\input{figs/latency}
\end{figure*}

\section{Results and Analysis}
\label{sec:results-and-analysis}

\paragraph{Efficiency-accuracy trade-off} 

We use SQuAD 1.1 to examine the effect of the proposed approach on the efficiency-accuracy trade-off.
When the underlying classifier was not trained with \ld{}, as proposed in this paper, the accuracy drops even more dramatically as more word vectors are dropped at each layer. The difference between standard transformer and \model{} is stark in Figure~\ref{fig:pareto_curve}. This verifies the importance of training a transformer in a way that makes it malleable for inference-time re-configuration. 

When the model was trained with the proposed \ld{}, we notice the efficacy of the proposed approach of using evolutionary search to find the optimal trade-off between inference efficiency and accuracy. The trade-off curve from the proposed search strategy has a larger area-under-curve (AUC) than when constant-rate length reduction was used to meet a target computational budget. It demonstrates the importance of using both \ld{} and evolutionary search. 

We make a minor observation that the proposed approach ends up with a significantly higher accuracy than DistillBERT when enough computational budget is allowed for inference ($\log \textrm{FLOPs} > 10$). This makes our approach desirable in a wide array of scenarios, as it does not require any additional pre-training stage, as does \db{}. With a severe constraint on the computational budget, the proposed approach could be used on \db{} to significantly improve the efficiency without compromising the accuracy.

\begin{figure*}[!t]
\centering
\begin{minipage}{0.72\textwidth}
\centering

\setlength{\tabcolsep}{0.5pt}

\footnotesize

\begin{tabular*}{0.98\textwidth}{l@{\extracolsep{\fill}}ccccccc}  
\toprule
\multicolumn{2}{c}{Model} & \multicolumn{2}{c}{SQuAD 1.1} & \multicolumn{2}{c}{MNLI-m} & \multicolumn{2}{c}{SST-2} \\
\makecell{Pretrained \\ Transformer} & Method & F1 & FLOPs & Acc & FLOPs & Acc & FLOPs \\
\midrule
\multirow{3}*{\bb} & Standard & 88.5 & 1.00x & 84.4 & 1.00x & 92.8 & 1.00x \\
                   & \lab     & 89.6 & 0.89x & 85.0 & 0.58x & 93.1 & 0.36x \\
                   & \las     & 88.7 & 0.45x & 84.4 & 0.35x & 92.8 & 0.35x \\
\midrule
\multirow{3}*{\db} & Standard & 85.8 & 1.00x & 80.9 & 1.00x & 90.6 & 1.00x \\
                   & \lab     & 86.3 & 0.81x & 81.5 & 0.56x & 92.0 & 0.55x \\
                   & \las     & 85.9 & 0.59x & 81.3 & 0.54x & 91.7 & 0.54x \\
\bottomrule
\end{tabular*}
\captionof{table}{
\label{tab:tasks}
Comparison results of standard Transformer and \model{}.
Among length configurations on the Pareto frontier of \model, we pick two representative points: \lab{} and \las{} as the most efficient one while having the highest accuracy and the accuracy higher than (or equal to) standard Transformer, respectively.
}
\end{minipage}

\quad
\input{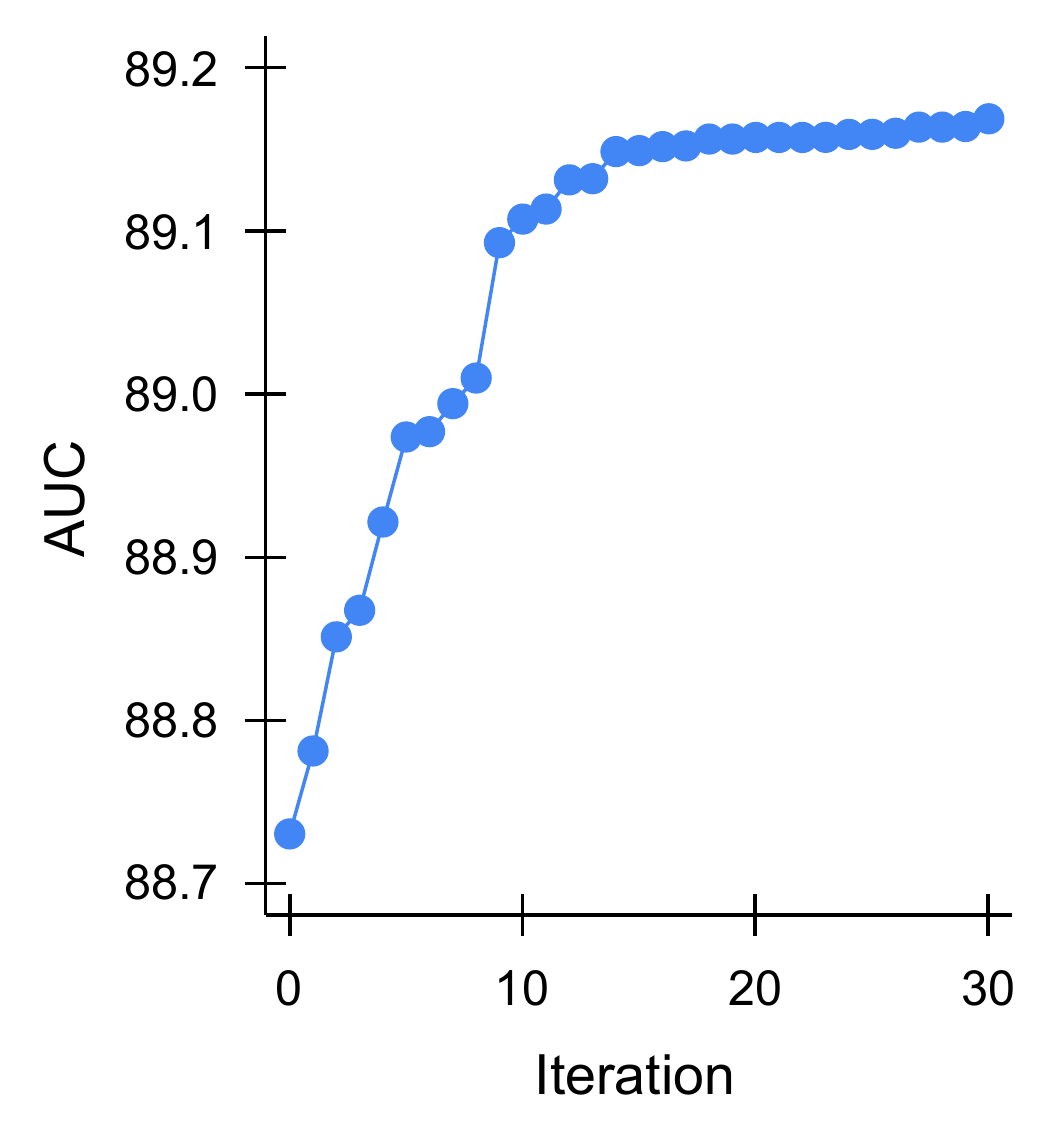}
\end{figure*}

\paragraph{Maximizing inference efficiency}

We consider all three tasks, SQuAD 1.1, MNLI-m, and SST-2, and investigate how much efficiency can be gained by the proposed approach with minimal sacrifice of accuracy. First, we look at how much efficiency could be gained without losing accuracy. That is, we use the length configuration that maximizes the inference efficiency (i.e., minimize the FLOPs) while ensuring that the accuracy is above or the same as the accuracy of the standard approach without any drop of word vectors. 
The results are presented in the rows marked with \las{} from Table~\ref{tab:tasks}. For example, in the case of \bb, the proposed approach reduces FLOPs by more than half across all three tasks.

From Figure~\ref{fig:pareto_curve}, we have observed that the proposed \model generalizes better than the standard, base model in some cases. Thus, we try to maximize both the inference efficiency and accuracy in order to see whether it is possible for the proposed algorithm to find a length configuration that both maximizes inference efficiency and improves accuracy. 
We present the results in the rows marked with \lab{} from Table~\ref{tab:tasks}.
For all cases, \model{} achieves higher accuracy than a standard transformer does while reducing FLOPs significantly.
Although it is not apparent from the table, tor MNLI-m and SST-2, the accuracy of the smallest sub-model is already greater than or equal to that of a standard transformer.

\paragraph{FLOPs vs. Latency}

As has been discussed in recent literature~(see, e.g., \citet{li2020train, chin2020towards}), the number of FLOPs is not a perfect indicator of the real latency measured in wall-clock time, as the latter is affected by the combination of hardware choice and network architecture. To understand the real-world impact of the proposed approach, we study the relationship between FLOPs, obtained by the proposed procedure, and wall-clock time measured on both CPU and GPU by measuring them while varying length configurations.
As shown in Figure~\ref{fig:latency}, FLOPs and latency exhibit near-linear correlation on GPU, when the minibatch size is $\geq 16$, and regardless of the minibatch size, on CPU.
In other words, the reduction in FLOPs with the proposed approach directly implies the reduction in wall-clock time.

\paragraph{Convergence of search}

Although the proposed approach is efficient in that it requires only one round of training, it needs a separate search stage for each target budget. It is important for evolutionary search to converge quickly in the number of forward sweeps of a validation set. As exemplified in Figure~\ref{fig:auc}, evolutionary search converges after about fifteen iterations.

\section{Comparison with Other Works}
Our framework allows a novel method for anytime prediction with adaptive sequence length given any transformers.
Thus, our goal is not state-of-the-art classification accuracy, although our experimental results (\textsection \ref{sec:results-and-analysis}) demonstrate that our method still attains a good accuracy level.

We emphasize that other adaptive computation works (\textsection \ref{sec:related_work}) are orthogonal with ours, meaning that various adaptive dimensions (sequence length, depth, attention head, hidden dimension, etc.) can be jointly used. 
In other words, even if other adaptive methods show better curves than ours, our method and theirs can boost each other when combined. 
We provide some comparison results with \pb{} (not anytime prediction method) and DynaBERT \cite{hou2020dynabert} (concurrent adaptive computation method) as follows.

\paragraph{Comparison with \pb{}}

According to \citet{goyal2020power}, PoWER-BERT achieves $2.6$x speedup for MNLI-m and $2.4$x speedup for SST-2 by losing $1\%$ of their accuracy.
\model{} obtains a $2.9$x speedup in terms of FLOPs without losing accuracy on MNLI-m and SST-2.
Considering Figure~\ref{fig:latency}, our speedup in execution time would be close to $2.9$x in the same setting of \pb{} where the time measurement is done with a batch size of $128$ on GPU.
It indicates that our model offers a better trade-off than \pb{}, even with a single model.

\paragraph{Comparison with DynaBERT}

According to \citet{hou2020dynabert}, DyanBERT obtains a gain of $+1.0$, $+0.1$, $+0.4$ for the best accuracy in SQuAD 1.1, MNLI-m, and SST-2, respectively, while \model{} achieves a gain of $+1.1$, $+0.6$, $+0.3$.
These results imply that \model{} can give a comparable (or better) performance with DynaBERT.  

\section{Related Work}
\label{sec:related_work}

The main purpose of the proposed algorithm is to improve the inference efficiency of a large-scale transformer. This goal has been pursued from various directions, and here we provide a brief overview of these earlier and some concurrent attempts in the context of the proposed approach.

\paragraph{Weight pruning} 

Weight pruning \citep{han2015deep} focuses on reducing the number of parameters that directly reflects the memory footprint of a model and indirectly correlates with inference speed. 
However, their actual speed-up in runtime is usually not significant, especially while executing a model with parallel computation using GPU devices \citep{tang2018flops,li2020train}.

\paragraph{Adaptive architecture}

There are three major axes along which computation can be reduced in a neural network; (1) input size/length, (2) network depth, and (3) network width.
The proposed approach, based on PoWER-BERT, adaptively reduces the input length as the input sequence is processed by the transformer layers. 
In our knowledge, \citet{goyal2020power} is the first work in this direction for transformers.
Funnel-Transformer \citep{dai2020funnel} and multi-scale transformer language models \citep{subramanian2020multi} also successfully reduce sequence length in the middle and rescale to full length for the final computation.
However, their inference complexity is fixed differently with \pb because they are not designed to control efficiency.
More recently, TR-BERT \citep{ye2021tr} introduces a policy network trained via reinforcement learning to decide which vectors to skip.

LayerDrop \citep{fan2019reducing} drops random layers during the training to be robust to pruning inspired by \citet{huang2016deep}. 
Word-level adaptive depth in \citet{elbayad2019depth} and \citet{liu2020explicitly} might seemingly resemble with length reduction,
but word vectors that reached the maximal layer are used for self-attention computation without updating themselves.
Escaping a network early \citep{teerapittayanon2016branchynet, huang2017multi} based on the confidence of the prediction \citep{xin2020deebert, xin2021berxit, schwartz2020right, liu2020fastbert, li2021accelerating} also offers a control over accuracy-efficiency trade-off by changing a threshold,
but it is difficult to tune a threshold for a desired computational budget because of the example-wise adaptive computation.

Slimmable neural networks \citep{yu2018slimmable, lee2018anytime} reduce the hidden dimension for the any-time prediction.
DynaBERT \citep{hou2020dynabert} can run at adaptive width (the number of attention heads and intermediate hidden dimension) and depth.
Hardware-aware Transformers \citep{wang2020hat} construct a design space with arbitrary encoder-decoder attention and heterogeneous layers in terms of different numbers of layers, attention heads, hidden dimension, and embedding dimension.
SpAtten \citep{wang2020spatten} performs cascade token and head pruning for an efficient algorithm-architecture co-design.

\paragraph{Structured dropout}

A major innovation we introduce over the existing PoWER-BERT is the use of stochastic, structured regularization to make a transformer robust to the choice of length configuration in the inference time.
\citet{rippel2014learning} proposes a nested dropout to learn ordered representations.
Similar to \ld, it samples an index from a prior distribution and drops all units with a larger index than the sampled one.

\paragraph{Search}

There have been a series of attempts at finding the optimal network configuration by solving a combinatorial optimization problem.
In computer vision, Once-for-All \citep{cai2019once} use an evolutionary search \citep{real2019regularized} to find a better configuration in dimensions of depth, width, kernel size, and resolution given computational budget.
Similarly but differently, our evolutionary search is \textit{mutli-objective} to find length configurations on the Pareto accuracy-efficiency frontier to cope with any possible computational budgets.
Moreover, we only change the sequence length of hidden vectors instead of architectural model size like dimensions.

\paragraph{Sequence Length}

Shortformer \citep{press2020shortformer} initially trained on shorter subsequences and then moved to longer ones achieves improved perplexity than a standard transformer with normal training while reducing overall training time.
Novel architectures with the efficient attention mechanism \cite{kitaev2020reformer, beltagy2020longformer, zaheer2020big, ainslie2020etc, choromanski2020rethinking, peng2021random} are suggested to reduce the transformer's quadratic computational complexity in the input sequence length.
\citet{tay2020efficient} and \citet{tay2020long} provide a survey of these efficient transformers and their benchmark comparison, respectively.

\section{Conclusion and Future Work}

In this work, we propose a new framework for training a transformer once and using it for efficient inference under any computational budget.
With the help of training with \ld{} and \dr{} process followed by the evolutionary search, our proposed \model{} allows any given transformer models to be used with any inference-time computational budget for both sequence-level and token-level classification tasks. Our experiments, on SQuAD 1.1, MNLI-m and SST-2, have revealed that the proposed algorithmic framework significantly pushes a better Pareto frontier on the trade-off between inference efficiency and accuracy. Furthermore, we have observed that the proposed \model{} could achieve up to 3x speed-up over the standard transformer without sacrificing accuracy, both in terms of FLOPs and wallclock time. 

Although our approach finds {\it an} optimal length configuration of a trained classifier per computational budget, it leaves a open question whether the proposed approach could be further extended to support per-instance length configuration by for instance training a small, auxiliary neural network for each computational budget. Yet another aspect we have not investigated in this paper is the applicability of the proposed approach to sequence generation, such as machine translation. We leave both of these research directions for the future.

Our approach is effective, as we have shown in this paper, and also quite simple to implement on top of existing language models.
We 
\ifaclfinal
release our implementation at \href{https://github.com/clovaai/length-adaptive-transformer}{https://github.com/clovaai/length-adaptive-transformer}, which is based on HuggingFace's \emph{Transformers} library \citep{wolf2019transformers},
\else
will release our implementation, which is based on HuggingFace's \emph{Transformers} library \citep{wolf2019transformers}, publicly 
\fi
and plan to adapt it for a broader set of transformer-based models and downstream tasks, including other modalities \citep{dosovitskiy2020image, touvron2020training, gulati2020conformer}.

\section*{Acknowledgments}
The authors appreciate Clova AI members and the anonymous reviewers for their constructive feedback.
Specifically, Dongyoon Han and Byeongho Heo introduced relevant works and gave insights from the view of the computer vision community.
We use Naver Smart Machine Learning \cite{sung2017nsml,kim2018nsml} platform for the experiments.


\bibliographystyle{acl_natbib}
\bibliography{ms}

\begin{thebibliography}{65}
\expandafter\ifx\csname natexlab\endcsname\relax\def\natexlab#1{#1}\fi

\bibitem[{Ainslie et~al.(2020)Ainslie, Ontan{\'o}n, Alberti, Cvicek, Fisher,
  Pham, Ravula, Sanghai, Wang, and Yang}]{ainslie2020etc}
Joshua Ainslie, Santiago Ontan{\'o}n, Chris Alberti, Vaclav Cvicek, Zachary
  Fisher, Philip Pham, Anirudh Ravula, Sumit Sanghai, Qifan Wang, and Li~Yang.
  2020.
\newblock Etc: Encoding long and structured inputs in transformers.
\newblock \emph{arXiv preprint arXiv:2004.08483}.

\bibitem[{Beltagy et~al.(2020)Beltagy, Peters, and
  Cohan}]{beltagy2020longformer}
Iz~Beltagy, Matthew~E Peters, and Arman Cohan. 2020.
\newblock Longformer: The long-document transformer.
\newblock \emph{arXiv preprint arXiv:2004.05150}.

\bibitem[{Brown et~al.(2020)Brown, Mann, Ryder, Subbiah, Kaplan, Dhariwal,
  Neelakantan, Shyam, Sastry, Askell et~al.}]{brown2020language}
Tom~B Brown, Benjamin Mann, Nick Ryder, Melanie Subbiah, Jared Kaplan, Prafulla
  Dhariwal, Arvind Neelakantan, Pranav Shyam, Girish Sastry, Amanda Askell,
  et~al. 2020.
\newblock Language models are few-shot learners.
\newblock \emph{arXiv preprint arXiv:2005.14165}.

\bibitem[{Cai et~al.(2019)Cai, Gan, and Han}]{cai2019once}
Han Cai, Chuang Gan, and Song Han. 2019.
\newblock Once for all: Train one network and specialize it for efficient
  deployment.
\newblock \emph{arXiv preprint arXiv:1908.09791}.

\bibitem[{Cao et~al.(2020)Cao, Balasubramanian, and
  Balasubramanian}]{cao2020towards}
Qingqing Cao, Aruna Balasubramanian, and Niranjan Balasubramanian. 2020.
\newblock Towards accurate and reliable energy measurement of nlp models.
\newblock \emph{arXiv preprint arXiv:2010.05248}.

\bibitem[{Chin et~al.(2020)Chin, Ding, Zhang, and Marculescu}]{chin2020towards}
Ting-Wu Chin, Ruizhou Ding, Cha Zhang, and Diana Marculescu. 2020.
\newblock Towards efficient model compression via learned global ranking.
\newblock In \emph{Proceedings of the IEEE/CVF Conference on Computer Vision
  and Pattern Recognition}, pages 1518--1528.

\bibitem[{Choromanski et~al.(2020)Choromanski, Likhosherstov, Dohan, Song,
  Gane, Sarlos, Hawkins, Davis, Mohiuddin, Kaiser
  et~al.}]{choromanski2020rethinking}
Krzysztof Choromanski, Valerii Likhosherstov, David Dohan, Xingyou Song,
  Andreea Gane, Tamas Sarlos, Peter Hawkins, Jared Davis, Afroz Mohiuddin,
  Lukasz Kaiser, et~al. 2020.
\newblock Rethinking attention with performers.
\newblock \emph{arXiv preprint arXiv:2009.14794}.

\bibitem[{Dai et~al.(2020)Dai, Lai, Yang, and Le}]{dai2020funnel}
Zihang Dai, Guokun Lai, Yiming Yang, and Quoc~V Le. 2020.
\newblock Funnel-transformer: Filtering out sequential redundancy for efficient
  language processing.
\newblock \emph{arXiv preprint arXiv:2006.03236}.

\bibitem[{Devlin et~al.(2018)Devlin, Chang, Lee, and
  Toutanova}]{devlin2018bert}
Jacob Devlin, Ming-Wei Chang, Kenton Lee, and Kristina Toutanova. 2018.
\newblock Bert: Pre-training of deep bidirectional transformers for language
  understanding.
\newblock \emph{arXiv preprint arXiv:1810.04805}.

\bibitem[{Dosovitskiy et~al.(2020)Dosovitskiy, Beyer, Kolesnikov, Weissenborn,
  Zhai, Unterthiner, Dehghani, Minderer, Heigold, Gelly
  et~al.}]{dosovitskiy2020image}
Alexey Dosovitskiy, Lucas Beyer, Alexander Kolesnikov, Dirk Weissenborn,
  Xiaohua Zhai, Thomas Unterthiner, Mostafa Dehghani, Matthias Minderer, Georg
  Heigold, Sylvain Gelly, et~al. 2020.
\newblock An image is worth 16x16 words: Transformers for image recognition at
  scale.
\newblock \emph{arXiv preprint arXiv:2010.11929}.

\bibitem[{Elbayad et~al.(2019)Elbayad, Gu, Grave, and Auli}]{elbayad2019depth}
Maha Elbayad, Jiatao Gu, Edouard Grave, and Michael Auli. 2019.
\newblock Depth-adaptive transformer.
\newblock \emph{arXiv preprint arXiv:1910.10073}.

\bibitem[{Fan et~al.(2019)Fan, Grave, and Joulin}]{fan2019reducing}
Angela Fan, Edouard Grave, and Armand Joulin. 2019.
\newblock Reducing transformer depth on demand with structured dropout.
\newblock \emph{arXiv preprint arXiv:1909.11556}.

\bibitem[{Goyal et~al.(2020)Goyal, Choudhury, Raje, Chakaravarthy, Sabharwal,
  and Verma}]{goyal2020power}
Saurabh Goyal, Anamitra~Roy Choudhury, Saurabh Raje, Venkatesan Chakaravarthy,
  Yogish Sabharwal, and Ashish Verma. 2020.
\newblock Power-bert: Accelerating bert inference via progressive word-vector
  elimination.
\newblock In \emph{International Conference on Machine Learning}, pages
  3690--3699. PMLR.

\bibitem[{Gulati et~al.(2020)Gulati, Qin, Chiu, Parmar, Zhang, Yu, Han, Wang,
  Zhang, Wu et~al.}]{gulati2020conformer}
Anmol Gulati, James Qin, Chung-Cheng Chiu, Niki Parmar, Yu~Zhang, Jiahui Yu,
  Wei Han, Shibo Wang, Zhengdong Zhang, Yonghui Wu, et~al. 2020.
\newblock Conformer: Convolution-augmented transformer for speech recognition.
\newblock \emph{arXiv preprint arXiv:2005.08100}.

\bibitem[{Han et~al.(2015)Han, Mao, and Dally}]{han2015deep}
Song Han, Huizi Mao, and William~J Dally. 2015.
\newblock Deep compression: Compressing deep neural networks with pruning,
  trained quantization and huffman coding.
\newblock \emph{arXiv preprint arXiv:1510.00149}.

\bibitem[{He et~al.(2020)He, Liu, Gao, and Chen}]{he2020deberta}
Pengcheng He, Xiaodong Liu, Jianfeng Gao, and Weizhu Chen. 2020.
\newblock Deberta: Decoding-enhanced bert with disentangled attention.
\newblock \emph{arXiv preprint arXiv:2006.03654}.

\bibitem[{Henderson et~al.(2020)Henderson, Hu, Romoff, Brunskill, Jurafsky, and
  Pineau}]{henderson2020towards}
Peter Henderson, Jieru Hu, Joshua Romoff, Emma Brunskill, Dan Jurafsky, and
  Joelle Pineau. 2020.
\newblock Towards the systematic reporting of the energy and carbon footprints
  of machine learning.
\newblock \emph{arXiv preprint arXiv:2002.05651}.

\bibitem[{Hinton et~al.(2015)Hinton, Vinyals, and Dean}]{hinton2015distilling}
Geoffrey Hinton, Oriol Vinyals, and Jeff Dean. 2015.
\newblock Distilling the knowledge in a neural network.
\newblock \emph{arXiv preprint arXiv:1503.02531}.

\bibitem[{Hou et~al.(2020)Hou, Shang, Jiang, and Liu}]{hou2020dynabert}
Lu~Hou, Lifeng Shang, Xin Jiang, and Qun Liu. 2020.
\newblock Dynabert: Dynamic bert with adaptive width and depth.
\newblock \emph{arXiv preprint arXiv:2004.04037}.

\bibitem[{Huang et~al.(2017)Huang, Chen, Li, Wu, van~der Maaten, and
  Weinberger}]{huang2017multi}
Gao Huang, Danlu Chen, Tianhong Li, Felix Wu, Laurens van~der Maaten, and
  Kilian~Q Weinberger. 2017.
\newblock Multi-scale dense networks for resource efficient image
  classification.
\newblock \emph{arXiv preprint arXiv:1703.09844}.

\bibitem[{Huang et~al.(2016)Huang, Sun, Liu, Sedra, and
  Weinberger}]{huang2016deep}
Gao Huang, Yu~Sun, Zhuang Liu, Daniel Sedra, and Kilian~Q Weinberger. 2016.
\newblock Deep networks with stochastic depth.
\newblock In \emph{European conference on computer vision}, pages 646--661.
  Springer.

\bibitem[{Kaplan et~al.(2020)Kaplan, McCandlish, Henighan, Brown, Chess, Child,
  Gray, Radford, Wu, and Amodei}]{kaplan2020scaling}
Jared Kaplan, Sam McCandlish, Tom Henighan, Tom~B Brown, Benjamin Chess, Rewon
  Child, Scott Gray, Alec Radford, Jeffrey Wu, and Dario Amodei. 2020.
\newblock Scaling laws for neural language models.
\newblock \emph{arXiv preprint arXiv:2001.08361}.

\bibitem[{Kim et~al.(2018)Kim, Kim, Seo, Kim, Park, Park, Jo, Kim, Yang, Kim
  et~al.}]{kim2018nsml}
Hanjoo Kim, Minkyu Kim, Dongjoo Seo, Jinwoong Kim, Heungseok Park, Soeun Park,
  Hyunwoo Jo, KyungHyun Kim, Youngil Yang, Youngkwan Kim, et~al. 2018.
\newblock Nsml: Meet the mlaas platform with a real-world case study.
\newblock \emph{arXiv preprint arXiv:1810.09957}.

\bibitem[{Kitaev et~al.(2020)Kitaev, Kaiser, and Levskaya}]{kitaev2020reformer}
Nikita Kitaev, {\L}ukasz Kaiser, and Anselm Levskaya. 2020.
\newblock Reformer: The efficient transformer.
\newblock \emph{arXiv preprint arXiv:2001.04451}.

\bibitem[{Lee and Shin(2018)}]{lee2018anytime}
Hankook Lee and Jinwoo Shin. 2018.
\newblock Anytime neural prediction via slicing networks vertically.
\newblock \emph{arXiv preprint arXiv:1807.02609}.

\bibitem[{Li et~al.(2021)Li, Shao, Sun, Yan, Qiu, and
  Huang}]{li2021accelerating}
Xiaonan Li, Yunfan Shao, Tianxiang Sun, Hang Yan, Xipeng Qiu, and Xuanjing
  Huang. 2021.
\newblock Accelerating bert inference for sequence labeling via early-exit.
\newblock \emph{arXiv preprint arXiv:2105.13878}.

\bibitem[{Li et~al.(2020)Li, Wallace, Shen, Lin, Keutzer, Klein, and
  Gonzalez}]{li2020train}
Zhuohan Li, Eric Wallace, Sheng Shen, Kevin Lin, Kurt Keutzer, Dan Klein, and
  Joseph~E Gonzalez. 2020.
\newblock Train large, then compress: Rethinking model size for efficient
  training and inference of transformers.
\newblock \emph{arXiv preprint arXiv:2002.11794}.

\bibitem[{Liu et~al.(2020{\natexlab{a}})Liu, Zhou, Zhao, Wang, Deng, and
  Ju}]{liu2020fastbert}
Weijie Liu, Peng Zhou, Zhe Zhao, Zhiruo Wang, Haotang Deng, and Qi~Ju.
  2020{\natexlab{a}}.
\newblock Fastbert: a self-distilling bert with adaptive inference time.
\newblock \emph{arXiv preprint arXiv:2004.02178}.

\bibitem[{Liu et~al.(2020{\natexlab{b}})Liu, Meng, Zhou, Chen, and
  Xu}]{liu2020explicitly}
Yijin Liu, Fandong Meng, Jie Zhou, Yufeng Chen, and Jinan Xu.
  2020{\natexlab{b}}.
\newblock Explicitly modeling adaptive depths for transformer.
\newblock \emph{arXiv preprint arXiv:2004.13542}.

\bibitem[{Michel et~al.(2019)Michel, Levy, and Neubig}]{michel2019sixteen}
Paul Michel, Omer Levy, and Graham Neubig. 2019.
\newblock Are sixteen heads really better than one?
\newblock In \emph{Advances in Neural Information Processing Systems}, pages
  14014--14024.

\bibitem[{Peng et~al.(2021)Peng, Pappas, Yogatama, Schwartz, Smith, and
  Kong}]{peng2021random}
Hao Peng, Nikolaos Pappas, Dani Yogatama, Roy Schwartz, Noah~A Smith, and
  Lingpeng Kong. 2021.
\newblock Random feature attention.
\newblock \emph{arXiv preprint arXiv:2103.02143}.

\bibitem[{Peters et~al.(2018)Peters, Neumann, Iyyer, Gardner, Clark, Lee, and
  Zettlemoyer}]{peters2018deep}
Matthew~E Peters, Mark Neumann, Mohit Iyyer, Matt Gardner, Christopher Clark,
  Kenton Lee, and Luke Zettlemoyer. 2018.
\newblock Deep contextualized word representations.
\newblock \emph{arXiv preprint arXiv:1802.05365}.

\bibitem[{Press et~al.(2020)Press, Smith, and Lewis}]{press2020shortformer}
Ofir Press, Noah~A Smith, and Mike Lewis. 2020.
\newblock Shortformer: Better language modeling using shorter inputs.
\newblock \emph{arXiv preprint arXiv:2012.15832}.

\bibitem[{Radford et~al.(2019)Radford, Wu, Child, Luan, Amodei, and
  Sutskever}]{radford2019language}
Alec Radford, Jeffrey Wu, Rewon Child, David Luan, Dario Amodei, and Ilya
  Sutskever. 2019.
\newblock Language models are unsupervised multitask learners.
\newblock \emph{OpenAI Blog}, 1(8):9.

\bibitem[{Raffel et~al.(2019)Raffel, Shazeer, Roberts, Lee, Narang, Matena,
  Zhou, Li, and Liu}]{raffel2019exploring}
Colin Raffel, Noam Shazeer, Adam Roberts, Katherine Lee, Sharan Narang, Michael
  Matena, Yanqi Zhou, Wei Li, and Peter~J Liu. 2019.
\newblock Exploring the limits of transfer learning with a unified text-to-text
  transformer.
\newblock \emph{arXiv preprint arXiv:1910.10683}.

\bibitem[{Rajpurkar et~al.(2016)Rajpurkar, Zhang, Lopyrev, and
  Liang}]{rajpurkar2016squad}
Pranav Rajpurkar, Jian Zhang, Konstantin Lopyrev, and Percy Liang. 2016.
\newblock Squad: 100,000+ questions for machine comprehension of text.
\newblock \emph{arXiv preprint arXiv:1606.05250}.

\bibitem[{Real et~al.(2019)Real, Aggarwal, Huang, and Le}]{real2019regularized}
Esteban Real, Alok Aggarwal, Yanping Huang, and Quoc~V Le. 2019.
\newblock Regularized evolution for image classifier architecture search.
\newblock In \emph{Proceedings of the aaai conference on artificial
  intelligence}, volume~33, pages 4780--4789.

\bibitem[{Rippel et~al.(2014)Rippel, Gelbart, and Adams}]{rippel2014learning}
Oren Rippel, Michael Gelbart, and Ryan Adams. 2014.
\newblock Learning ordered representations with nested dropout.
\newblock In \emph{International Conference on Machine Learning}, pages
  1746--1754.

\bibitem[{Sanh et~al.(2019)Sanh, Debut, Chaumond, and
  Wolf}]{sanh2019distilbert}
Victor Sanh, Lysandre Debut, Julien Chaumond, and Thomas Wolf. 2019.
\newblock Distilbert, a distilled version of bert: smaller, faster, cheaper and
  lighter.
\newblock \emph{arXiv preprint arXiv:1910.01108}.

\bibitem[{Schwartz et~al.(2019)Schwartz, Dodge, Smith, and
  Etzioni}]{schwartz2019green}
Roy Schwartz, Jesse Dodge, Noah~A Smith, and Oren Etzioni. 2019.
\newblock Green ai.
\newblock \emph{arXiv preprint arXiv:1907.10597}.

\bibitem[{Schwartz et~al.(2020)Schwartz, Stanovsky, Swayamdipta, Dodge, and
  Smith}]{schwartz2020right}
Roy Schwartz, Gabi Stanovsky, Swabha Swayamdipta, Jesse Dodge, and Noah~A
  Smith. 2020.
\newblock The right tool for the job: Matching model and instance complexities.
\newblock \emph{arXiv preprint arXiv:2004.07453}.

\bibitem[{See et~al.(2016)See, Luong, and Manning}]{see2016compression}
Abigail See, Minh-Thang Luong, and Christopher~D Manning. 2016.
\newblock Compression of neural machine translation models via pruning.
\newblock \emph{arXiv preprint arXiv:1606.09274}.

\bibitem[{Shoeybi et~al.(2019)Shoeybi, Patwary, Puri, LeGresley, Casper, and
  Catanzaro}]{shoeybi2019megatron}
Mohammad Shoeybi, Mostofa Patwary, Raul Puri, Patrick LeGresley, Jared Casper,
  and Bryan Catanzaro. 2019.
\newblock Megatron-lm: Training multi-billion parameter language models using
  gpu model parallelism.
\newblock \emph{arXiv preprint arXiv:1909.08053}.

\bibitem[{Srivastava et~al.(2014)Srivastava, Hinton, Krizhevsky, Sutskever, and
  Salakhutdinov}]{srivastava2014dropout}
Nitish Srivastava, Geoffrey Hinton, Alex Krizhevsky, Ilya Sutskever, and Ruslan
  Salakhutdinov. 2014.
\newblock Dropout: a simple way to prevent neural networks from overfitting.
\newblock \emph{The journal of machine learning research}, 15(1):1929--1958.

\bibitem[{Strubell et~al.(2019)Strubell, Ganesh, and
  McCallum}]{strubell2019energy}
Emma Strubell, Ananya Ganesh, and Andrew McCallum. 2019.
\newblock Energy and policy considerations for deep learning in nlp.
\newblock \emph{arXiv preprint arXiv:1906.02243}.

\bibitem[{Subramanian et~al.(2020)Subramanian, Collobert, Ranzato, and
  Boureau}]{subramanian2020multi}
Sandeep Subramanian, Ronan Collobert, Marc'Aurelio Ranzato, and Y-Lan Boureau.
  2020.
\newblock Multi-scale transformer language models.
\newblock \emph{arXiv preprint arXiv:2005.00581}.

\bibitem[{Sun et~al.(2019)Sun, Cheng, Gan, and Liu}]{sun2019patient}
Siqi Sun, Yu~Cheng, Zhe Gan, and Jingjing Liu. 2019.
\newblock Patient knowledge distillation for bert model compression.
\newblock \emph{arXiv preprint arXiv:1908.09355}.

\bibitem[{Sung et~al.(2017)Sung, Kim, Jo, Yang, Kim, Lausen, Kim, Lee, Kwak, Ha
  et~al.}]{sung2017nsml}
Nako Sung, Minkyu Kim, Hyunwoo Jo, Youngil Yang, Jingwoong Kim, Leonard Lausen,
  Youngkwan Kim, Gayoung Lee, Donghyun Kwak, Jung-Woo Ha, et~al. 2017.
\newblock Nsml: A machine learning platform that enables you to focus on your
  models.
\newblock \emph{arXiv preprint arXiv:1712.05902}.

\bibitem[{Tang et~al.(2018)Tang, Adhikari, and Lin}]{tang2018flops}
Raphael Tang, Ashutosh Adhikari, and Jimmy Lin. 2018.
\newblock Flops as a direct optimization objective for learning sparse neural
  networks.
\newblock \emph{arXiv preprint arXiv:1811.03060}.

\bibitem[{Tay et~al.(2020{\natexlab{a}})Tay, Dehghani, Abnar, Shen, Bahri,
  Pham, Rao, Yang, Ruder, and Metzler}]{tay2020long}
Yi~Tay, Mostafa Dehghani, Samira Abnar, Yikang Shen, Dara Bahri, Philip Pham,
  Jinfeng Rao, Liu Yang, Sebastian Ruder, and Donald Metzler.
  2020{\natexlab{a}}.
\newblock Long range arena: A benchmark for efficient transformers.
\newblock \emph{arXiv preprint arXiv:2011.04006}.

\bibitem[{Tay et~al.(2020{\natexlab{b}})Tay, Dehghani, Bahri, and
  Metzler}]{tay2020efficient}
Yi~Tay, Mostafa Dehghani, Dara Bahri, and Donald Metzler. 2020{\natexlab{b}}.
\newblock Efficient transformers: A survey.
\newblock \emph{arXiv preprint arXiv:2009.06732}.

\bibitem[{Teerapittayanon et~al.(2016)Teerapittayanon, McDanel, and
  Kung}]{teerapittayanon2016branchynet}
Surat Teerapittayanon, Bradley McDanel, and Hsiang-Tsung Kung. 2016.
\newblock Branchynet: Fast inference via early exiting from deep neural
  networks.
\newblock In \emph{2016 23rd International Conference on Pattern Recognition
  (ICPR)}, pages 2464--2469. IEEE.

\bibitem[{Touvron et~al.(2020)Touvron, Cord, Douze, Massa, Sablayrolles, and
  J{\'e}gou}]{touvron2020training}
Hugo Touvron, Matthieu Cord, Matthijs Douze, Francisco Massa, Alexandre
  Sablayrolles, and Herv{\'e} J{\'e}gou. 2020.
\newblock Training data-efficient image transformers \& distillation through
  attention.
\newblock \emph{arXiv preprint arXiv:2012.12877}.

\bibitem[{Vaswani et~al.(2017)Vaswani, Shazeer, Parmar, Uszkoreit, Jones,
  Gomez, Kaiser, and Polosukhin}]{vaswani2017attention}
Ashish Vaswani, Noam Shazeer, Niki Parmar, Jakob Uszkoreit, Llion Jones,
  Aidan~N Gomez, {\L}ukasz Kaiser, and Illia Polosukhin. 2017.
\newblock Attention is all you need.
\newblock In \emph{Advances in neural information processing systems}, pages
  5998--6008.

\bibitem[{Wang et~al.(2018)Wang, Singh, Michael, Hill, Levy, and
  Bowman}]{wang2018glue}
Alex Wang, Amanpreet Singh, Julian Michael, Felix Hill, Omer Levy, and Samuel~R
  Bowman. 2018.
\newblock Glue: A multi-task benchmark and analysis platform for natural
  language understanding.
\newblock \emph{arXiv preprint arXiv:1804.07461}.

\bibitem[{Wang et~al.(2020{\natexlab{a}})Wang, Wu, Liu, Cai, Zhu, Gan, and
  Han}]{wang2020hat}
Hanrui Wang, Zhanghao Wu, Zhijian Liu, Han Cai, Ligeng Zhu, Chuang Gan, and
  Song Han. 2020{\natexlab{a}}.
\newblock Hat: Hardware-aware transformers for efficient natural language
  processing.
\newblock \emph{arXiv preprint arXiv:2005.14187}.

\bibitem[{Wang et~al.(2020{\natexlab{b}})Wang, Zhang, and
  Han}]{wang2020spatten}
Hanrui Wang, Zhekai Zhang, and Song Han. 2020{\natexlab{b}}.
\newblock Spatten: Efficient sparse attention architecture with cascade token
  and head pruning.
\newblock \emph{arXiv preprint arXiv:2012.09852}.

\bibitem[{Wolf et~al.(2019)Wolf, Debut, Sanh, Chaumond, Delangue, Moi, Cistac,
  Rault, Louf, Funtowicz et~al.}]{wolf2019transformers}
Thomas Wolf, Lysandre Debut, Victor Sanh, Julien Chaumond, Clement Delangue,
  Anthony Moi, Pierric Cistac, Tim Rault, R{\'e}mi Louf, Morgan Funtowicz,
  et~al. 2019.
\newblock Transformers: State-of-the-art natural language processing.
\newblock \emph{arXiv preprint arXiv:1910.03771}.

\bibitem[{Xin et~al.(2020)Xin, Tang, Lee, Yu, and Lin}]{xin2020deebert}
Ji~Xin, Raphael Tang, Jaejun Lee, Yaoliang Yu, and Jimmy Lin. 2020.
\newblock Deebert: Dynamic early exiting for accelerating bert inference.
\newblock \emph{arXiv preprint arXiv:2004.12993}.

\bibitem[{Xin et~al.(2021)Xin, Tang, Yu, and Lin}]{xin2021berxit}
Ji~Xin, Raphael Tang, Yaoliang Yu, and Jimmy Lin. 2021.
\newblock Berxit: Early exiting for bert with better fine-tuning and extension
  to regression.
\newblock In \emph{Proceedings of the 16th Conference of the European Chapter
  of the Association for Computational Linguistics: Main Volume}, pages
  91--104.

\bibitem[{Yang et~al.(2019)Yang, Dai, Yang, Carbonell, Salakhutdinov, and
  Le}]{yang2019xlnet}
Zhilin Yang, Zihang Dai, Yiming Yang, Jaime Carbonell, Russ~R Salakhutdinov,
  and Quoc~V Le. 2019.
\newblock Xlnet: Generalized autoregressive pretraining for language
  understanding.
\newblock In \emph{Advances in neural information processing systems}, pages
  5754--5764.

\bibitem[{Ye et~al.(2021)Ye, Lin, Huang, and Sun}]{ye2021tr}
Deming Ye, Yankai Lin, Yufei Huang, and Maosong Sun. 2021.
\newblock Tr-bert: Dynamic token reduction for accelerating bert inference.
\newblock \emph{arXiv preprint arXiv:2105.11618}.

\bibitem[{Yu and Huang(2019)}]{yu2019universally}
Jiahui Yu and Thomas~S Huang. 2019.
\newblock Universally slimmable networks and improved training techniques.
\newblock In \emph{Proceedings of the IEEE International Conference on Computer
  Vision}, pages 1803--1811.

\bibitem[{Yu et~al.(2018)Yu, Yang, Xu, Yang, and Huang}]{yu2018slimmable}
Jiahui Yu, Linjie Yang, Ning Xu, Jianchao Yang, and Thomas Huang. 2018.
\newblock Slimmable neural networks.
\newblock \emph{arXiv preprint arXiv:1812.08928}.

\bibitem[{Zaheer et~al.(2020)Zaheer, Guruganesh, Dubey, Ainslie, Alberti,
  Ontanon, Pham, Ravula, Wang, Yang et~al.}]{zaheer2020big}
Manzil Zaheer, Guru Guruganesh, Avinava Dubey, Joshua Ainslie, Chris Alberti,
  Santiago Ontanon, Philip Pham, Anirudh Ravula, Qifan Wang, Li~Yang, et~al.
  2020.
\newblock Big bird: Transformers for longer sequences.
\newblock \emph{arXiv preprint arXiv:2007.14062}.

\end{thebibliography}


\end{document}